\documentclass{article}
\usepackage{spconf,amsmath,graphicx}

\usepackage{times}
\usepackage{epsfig}
\usepackage{graphicx}
\usepackage{amssymb}
\usepackage{booktabs}
\usepackage[T1]{fontenc}
\usepackage{amsmath}
\usepackage{booktabs}
\usepackage{dsfont}
\usepackage{etoolbox}
\usepackage{graphicx}
\usepackage{makecell}
\usepackage{multicol}
\usepackage{multirow}
\usepackage{nicefrac}
\usepackage{paralist}
\usepackage{pifont}
\usepackage{siunitx}
\usepackage{xspace}
\usepackage{placeins}
\usepackage{floatrow}
\usepackage{hyperref}
\usepackage{fdsymbol}

\usepackage{enumitem}
\setlist{nosep, leftmargin=14pt}

\usepackage{mwe} % to get dummy images

\newcommand{\modelname}{DISPR\xspace}

\newcommand{\reals}{\ensuremath{\mathds{R}}\xspace}

\title{A Diffusion Model Predicts 3D Shapes from 2D Microscopy Images}

\name{
    \small
    Dominik J.\,E.\ Waibel$^{1,3,\star}$ 
    \qquad 
    Ernst Röell$^{1,2}$ 
    \qquad 
    Bastian Rieck$^{1,2,\dag}$
    \qquad
    Raja Giryes$^{4,\dag}$
    \qquad
    Carsten Marr$^{1,\dag}$
    }

\address{
    \small\vspace{-.1cm}
    $^{1}$ Institute of AI for Health, Helmholtz Munich -- German Research Centre for Environmental Health, Neuherberg, Germany \\
    \small\vspace{-.1cm}
    $^{2}$  TUM School of Computation, Information and Technology, Technical University of Munich, Munich Germany\\
    \small\vspace{-.1cm}
    $^{3}$ TUM School of Life Sciences, Technical University of Munich, Munich, Germany \\
    \small\vspace{-.1cm}
    $^{4}$ Faculty of Engineering, Tel Aviv University, Tel Aviv, Israel \\
    \small\vspace{-.1cm} 
    $^{\star}$ Current affiliation: AstraZeneca Computational Pathology, Oncology R\&D, Munich, Germany\\
    \small\vspace{-.1cm}
    $^{\dag}$ These authors share corresponding authorship
    }

\begin{document}
\ninept
\maketitle
\begin{abstract}
Diffusion models are a special type of generative model, capable of synthesising new data from a learnt distribution. 
We introduce \modelname, a diffusion-based model for solving the inverse problem of three-dimensional (3D) cell shape prediction from two-dimensional (2D) single cell microscopy images. 
Using the 2D microscopy image as a prior, \modelname\ is conditioned to predict realistic  3D shape reconstructions.
%Since diffusion models rely on a stochastic sampling process that allows to generate infinitely many predictions, we exploit this process to generate five reconstructions of every microscopy image.
To showcase the applicability of \modelname\ as a data augmentation tool in a feature-based single cell classification task, we extract morphological features from the red blood cells grouped into six highly imbalanced classes. 
Adding features from the \modelname\ predictions to the three minority classes improved the macro F1 score from $F1_\text{macro} = 55.2 \pm 4.6\%$ to $F1_\text{macro} = 72.2 \pm 4.9\%$.  
We thus demonstrate that diffusion models can be successfully applied to inverse biomedical problems, and that they learn to reconstruct 3D shapes with realistic morphological features from 2D microscopy images.
\end{abstract}

\section{Introduction}
\label{sec:intro}

% \begin{figure}[t]
% \begin{center}
%     \fbox{%\rule{0pt}{2in} 
%         \rule{0.01\linewidth}{0pt}
%         \includegraphics[width=0.9\linewidth]{Figures/WACV_DiffusionSHAPR_Figures.006.png}}
% \end{center}
%     \caption{Diffusion steps $t$ of one forward pass through \modelname\ visualized during inference between step $t=999$ with the largest amount of Gaussian noise and the \modelname\ prediction of $t=0$, visualized for a stromatocyte cell (see also Section \ref{sec:Methods}). Note that the zoom level in the visualizations is not uniform.}
%     \label{fig:Diffusion_steps}
% \end{figure}

Diffusion models are a class of generative models, showing superior performance as compared to other generative models in creating realistic images when trained on natural image datasets. 
We apply a diffusion model to the reconstruction of three-dimensional single cell shapes~(3D) from two-dimensional~(2D) microscopy images, using the 2D image as a prior.

%\begin{figure*}[t]
%\begin{center}
%    \fbox{%\rule{0pt}{2in} 
%        \rule{0.01\linewidth}{0pt}
%        \includegraphics[width=0.7\linewidth]{Figures/WACV_DiffusionSHAPR_Figures.001.png}}
%\end{center}
%\begin{figure}[t]
%    \fbox{%\rule{0pt}{2in} 
%        \rule{0.01\linewidth}{0pt}
%        \includegraphics[width=\linewidth]{Figures/DiffusionSHAPR_Figures.001.png}}
%\caption{\modelname\ is trained to denoise the 3D volume $\mathbf{x}_{b,t}$ containing added stochastic Gaussian noise, to obtain $\mathbf{x}_{b,t-1}$, thereby reversing the noising step $q$. In each forward pass $p_{\theta}$, we constrain our 3D model with one of the 2D images $b$. During inference, the forward pass $p_{\theta}$ is repeated $T$ times to obtain the prediction $\mathbf{x}_{b,0}$ (see Section \ref{sec:Methods}).}
%\label{fig:Diffusion_SHAPR}
%\end{figure*}

\begin{figure}[htb]
% \begin{minipage}[b]{1.0\linewidth}
  \centering
  \centerline{\includegraphics[width=\linewidth]{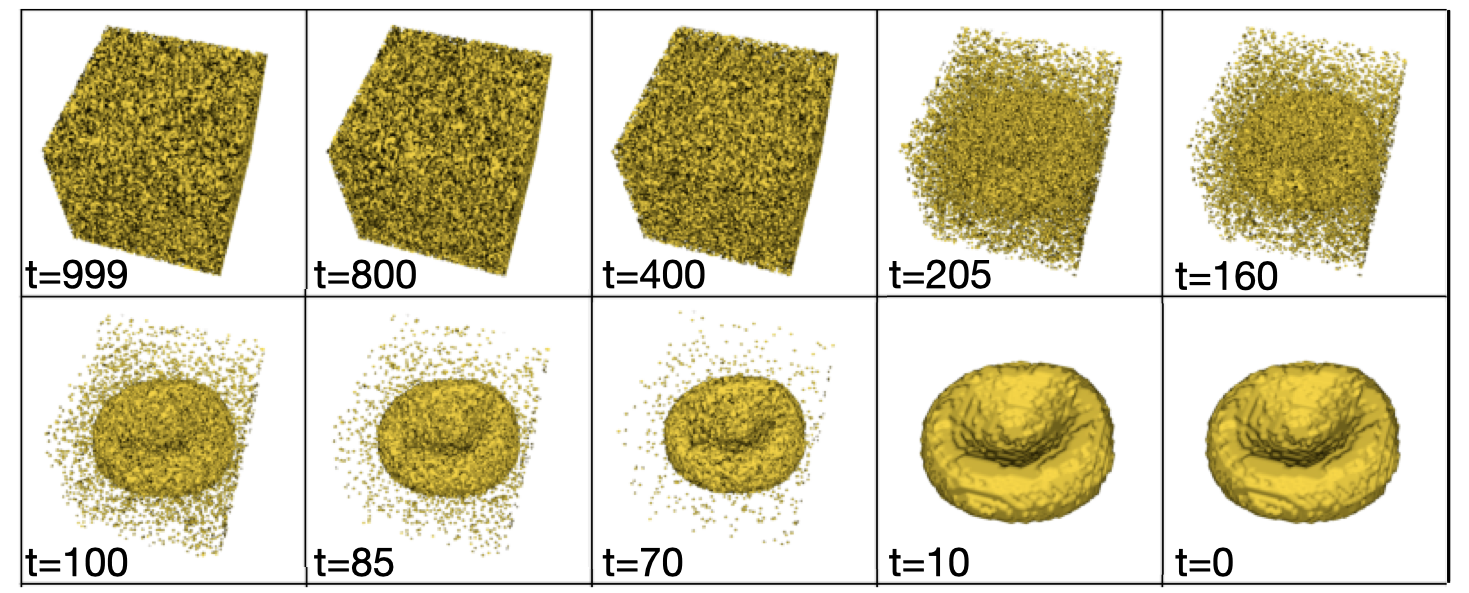}}
%  \vspace{2.0cm}
    \caption{Diffusion steps of one forward pass through \modelname\ visualized during inference between step $t=999$ with the largest amount of Gaussian noise and the \modelname\ prediction of $t=0$, for a stromatocyte cell (see section \ref{sec:Methods}). Note that the zoom level in the visualizations is not uniform.}
    \label{fig:Diffusion_SHAPR}
% \end{minipage}
\end{figure}

Our work is motivated by the observation that imaging speed is a central limitation for microscopic confocal imaging of single cells. 
This is due to the fact that the sequential imaging and stacking of slices of 2D images is time-consuming and toxic for cells. 
However, the benefits of 3D microscopy are staggering: assessing morphological information of individual cells---such as their volume, shape or surface---via 3D microscopy promises new insights into blood disorders such as sickle cell anemia \cite{diez2010shape}. Imaging a large number of cells at high resolution in 3D is rather costly.
% Ideally however, one would extrapolate the 3D shape of a single cell from an individual 2D microscopy image, thus enabling the assessment of morphological features based on sparse data.
Therefore, researchers have to find ways how they can optimally balance throughput and resolution of 3D microscopy data.

% \begin{figure}[t]
% \begin{center}
%     \fbox{%\rule{0pt}{2in} 
%         \rule{0.01\linewidth}{0pt}
%         \includegraphics[width=0.9\linewidth]{Figures/WACV_DiffusionSHAPR_Figures.006.png}}
% \end{center}
%     \caption{Diffusion steps $t$ of one forward pass through \modelname\ visualized during inference between step $t=999$ with the largest amount of Gaussian noise and the \modelname\ prediction of $t=0$, visualized for a stromatocyte cell (see also Section \ref{sec:Methods}). Note that the zoom level in the visualizations is not uniform.}
%     \label{fig:Diffusion_steps}
% \end{figure}

\begin{figure}[htb]
%\begin{minipage}[b]{1.0\linewidth}
  \centering
  \centerline{\includegraphics[width=8.0cm]{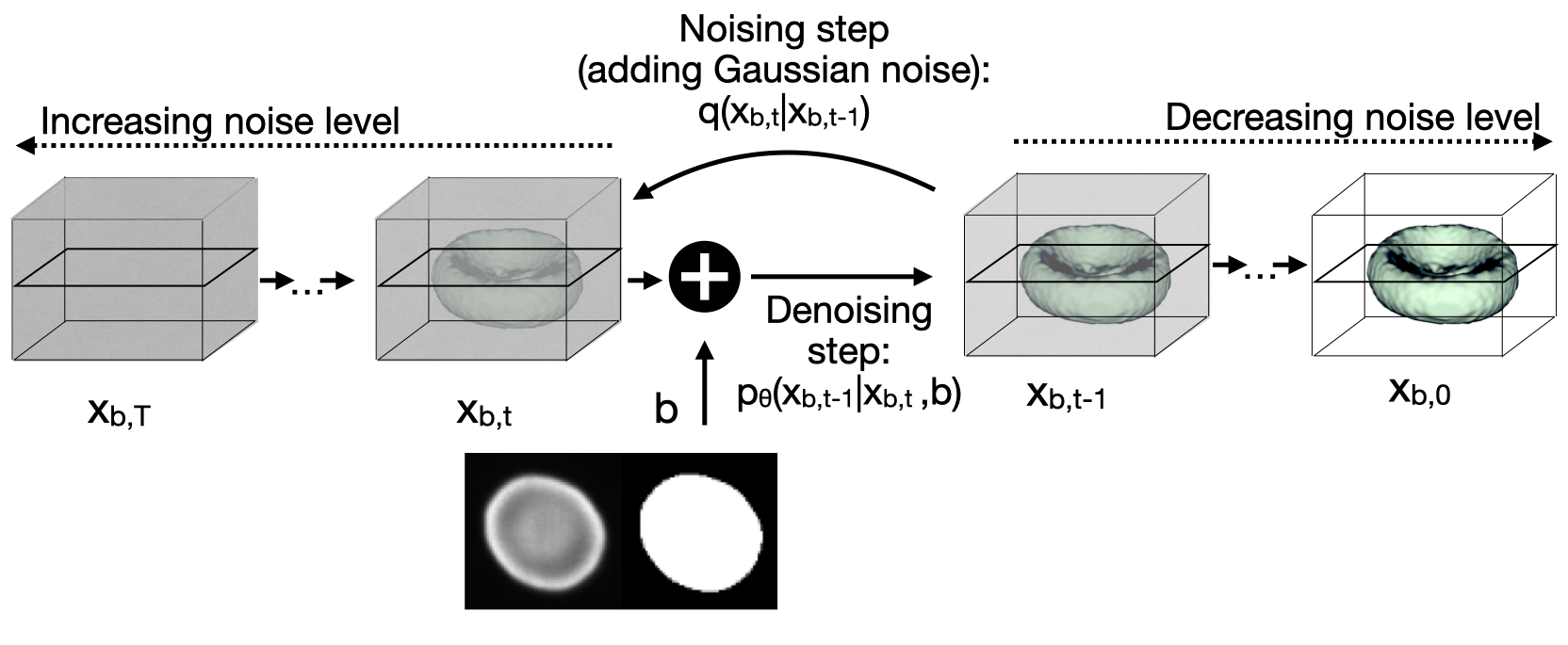}}
%  \vspace{2.0cm}
  \caption{\modelname\ is trained to denoise the 3D volume $\mathbf{x}_{b,t}$ containing added stochastic Gaussian noise, to obtain $\mathbf{x}_{b,t-1}$, thus reversing the noising step $q$. In each forward pass $p_{\theta}$, we constrain our 3D model with one 2D image $b$ containing a fluorecent image and a mask. During inference, the forward pass $p_{\theta}$ is repeated $T$ times to obtain the prediction $\mathbf{x}_{b,0}$~(see section \ref{sec:Methods}).}
  \label{fig:Diffusion_steps}
%\end{minipage}
\end{figure}

% \begin{figure}[t]
%    \fbox{%\rule{0pt}{2in} 
%        %\rule{0.01\linewidth}{0pt}
%        \includegraphics[width=\linewidth]{Figures/WACV_DiffusionSHAPR_Figures.008.png}}
%    \caption{Red blood cells exhibit similar morphological features between the groundtruth (3rd column, %yellow background) and the predictions of \modelname\ (white background). 
%    The 2D microscopy image (with a fluorescence and segmented channel, dark background) is used as the model's input to predict the 3D shapes. Spherocytes, stomatocytes, discocytes, and echinocytes are combined into the SDE class, which is characterized by the stomatocyte–discocyte–echinocyte transformation \cite{Simionato21a}.}
%    \label{fig:Visualization}
%\end{figure}

%Multiple microscopy techniques that reveal 3D information have recently been developed, including optical diffraction tomography~\cite{Sung2009-mw}, digital holographic imaging~\cite{anand2017automated}, and integral imaging microscopy~\cite{Javidi2006-ry,Martinez-Corral2018-yh}.
%
A second challenge in the medical domain, especially in 3D microscopy, is the manual labor involved in labeling, segmenting and annotating data samples. This dependence makes it both time consuming and expensive. Moreover, datasets are most likely imbalanced, further impeding the application of deep learning methods.
Data augmentation techniques generate synthetic training samples from the training dataset, adding to the total volume of training data available for deep learning models.
The distributional nature of our model lends itself particularly well for data augmentation. 
%After training, our model is capable of generating synthetic samples to augment underrepresented classes.
We demonstrate this in a red blood cell classification task, where we augment the training set with synthetic samples and train a classifier.

% Our approach allows us to generate a distribution of plausible reconstructions that we can sample from.
% We thus  constrain our diffusion-based model to predict 3D cell shapes that are realistic reconstructions of this cell imaged in 2D, acknowledging the ambiguity of the inverse problem, as our objective is not to obtain a single, deterministic reconstruction, but training a distribution of plausible reconstructions.

Existing approaches \cite{Waibel2021-or, Waibel2022-ng} aim to predict a single, deterministic reconstruction of a biological structure for each single 2D input. 
Others focus on predicting the shapes of natural objects such as air planes, cars, and furniture from photographs, creating either meshes~\cite{Gkioxari2019-ba,Wang2018-qg}, voxel volumes~\cite{Choy2016-fa,watson2022novel}, or point clouds~\cite{Fan2017-uf}.
%
% Fluorescence microscopy imaging, where cellular structures are labelled with a fluorescent stain, is fundamentally different from real-world photographs in terms of color, contrast, and object orientation~\cite{Waibel2021-or}, requiring tailored models that are capable of leveraging dosmain-specific information.
In the biomedical domain, SHAPR, a deterministic autoencoder, was developed by Waibel et al.\ to predict the shape and morphology of individual mammalian cells from 2D microscopy images~\cite{Waibel2021-or}. 
SHAPR has recently been refined with a topological loss function~\cite{Waibel2022-ng}. 
Both methods are based on an autoencoder that predicts one shape for each input image.
By contrast, our method can be used to predict a distribution of 3D shape reconstructions for each input image.

\begin{figure}[tb]
%\begin{minipage}[b]{1.0\linewidth}
  \centering
  \centerline{\includegraphics[width=8.0cm]{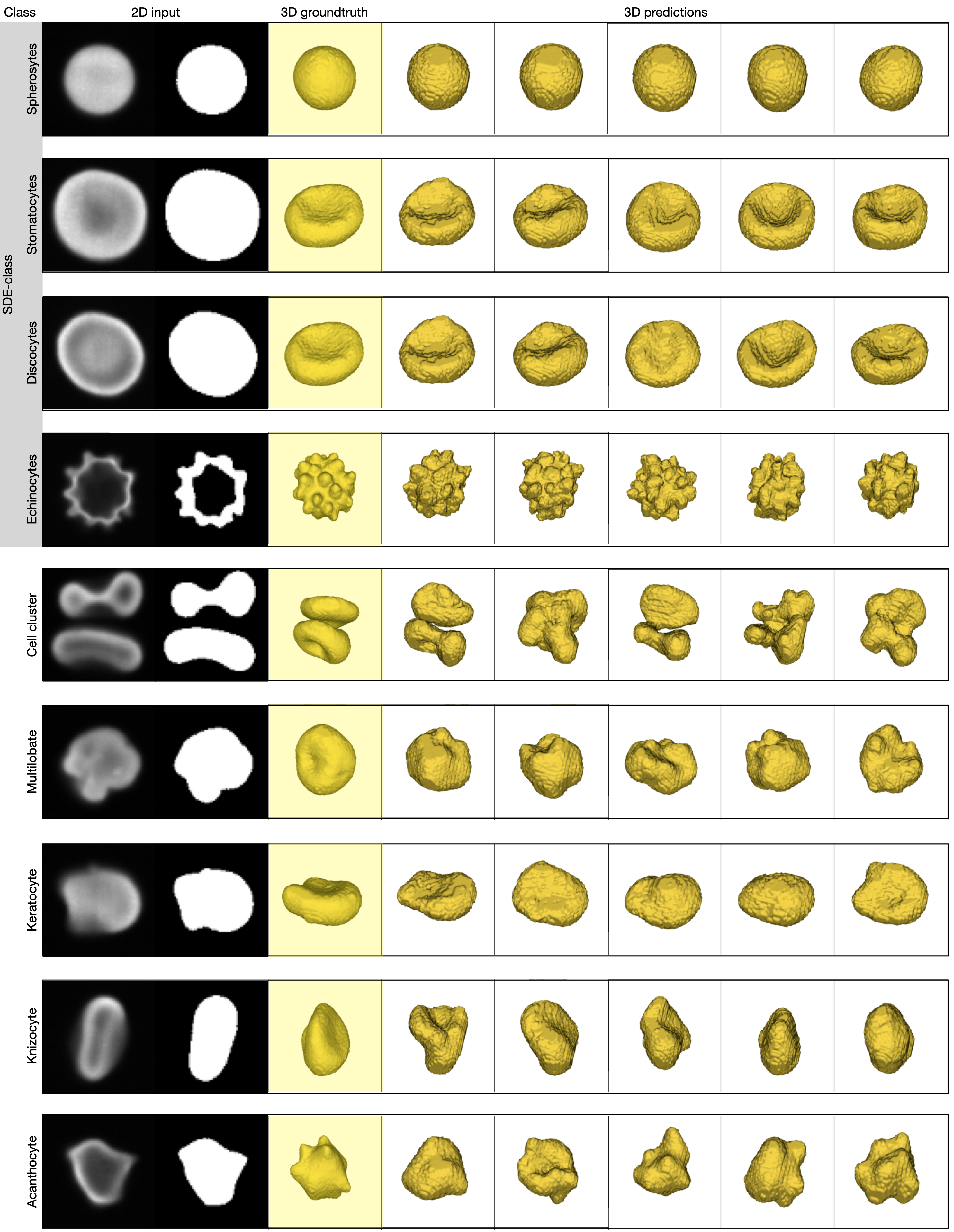}}
%  \vspace{2.0cm}
  \caption{Red blood cells exhibit similar morphological features between groundtruth (3rd column, yellow background) and \modelname\ predictions (white background). 
    A 2D microscopy image (with a fluorescence and segmented channel, dark background) is used as the model's input to predict the 3D shapes. 
    %Spherocytes, stomatocytes, discocytes, and echinocytes are combined into the SDE class, which is characterized by the stomatocyte–discocyte–echinocyte transformation \cite{Simionato21a}.
    }
    \label{fig:Visualization}
%\end{minipage}
\end{figure}

Diffusion models have gained much attention due to their astonishing performance in generating realistically-looking images~\cite{Saharia2022-xs,Nichol2021-ga}. 
Denoising diffusion implicit models \cite{Song2020-zk} have substantially improved the noise sampling scheme by skipping multiple noise sampling steps. 
Denoising diffusion probabilistic models were further improved in their loss function by architecture changes and by classifier guidance during sampling, leading to improved image quality of the predictions~\cite{Nichol2021-ga,Dhariwal_undated-xt}. While most diffusion models are applied in the natural image domain, Wolleb et al.\ have used them for segmentation of MRI images~\cite{Wolleb2021-xk} and anomaly detection~\cite{Wolleb2022-sq} in multimodal brain images, showing their applicability in the medical domain for segmentation of 2D MRI images. Diffusion models have been used of 2D MRI and CT image synthesis \cite{khader2022medical, ozbey2022unsupervised,ali2022spot}, to synthesise longitudinal MRI images \cite{yoon2022sadm} and 4D MRI images \cite{kim2022diffusion}.  

We go beyond previous work and propose a diffusion model, hereafter called \textbf{DI}ffusion based \textbf{S}hape \textbf{PR}ediction, \modelname\footnote{See https://github.com/marrlab/DISPR.} 
for predicting single cell shapes that are realistic 3D reconstructions from 2D microscopy images. 
% We thus describe the first application of diffusion models to the challenging inverse problem of shape reconstruction.

\section{Methods}
\label{sec:Methods}
We first provide a brief overview of diffusion models in general before detailing the specific architectural choices of \modelname, our model for predicting 3D cell shapes from 2D microscopy images.

A diffusion model contains two processes, a forward diffusion process
and a reconstruction (or denoising) process. 
In the forward diffusion process, Gaussian noise is successively added to an input image in a 
predefined number of steps, generating a sequence of images with increasing noise. This sequence will in the limit approach a 
normal distribution with unit variance, losing all information. The sequences of noisy images will then serve as the 
training data for the reconstruction process.
The diffusion model is trained to recover the original image given the sequence of noisy images, in essence denoising them.

To describe the forward diffusion process, let $\mathbf{x}_0$ denote the original 3D image.
We denote the normal distribution with mean $\mu$ and variance $\beta$ applied to $\mathbf{x}_t$ by 
$\mathcal{N}(\mathbf{x}_t;\mu,\beta)$.
The forward diffusion of $\mathbf{x}_{t-1}$ to $\mathbf{x}_{t}$ is recursively given by 
\begin{equation}
q(\mathbf{x}_t \vert \mathbf{x}_{t-1}) = \mathcal{N}(\mathbf{x}_t; \sqrt{1 - \beta_t} \mathbf{x}_{t-1}, \beta_t\mathbf{I}),
\label{eqn:transition}
\end{equation}
where $\mathbf{I}$ denotes the identity matrix. Note that we allow the variance $\beta_t$ to increase with each step~\cite{Ho_undated-ak,Nichol2021-ga}. The forward diffusion process 
is repeated for a fixed, predefined number of steps $T$. In our case, in line with the literature, we set $T=1000$.
Repeated application of Eq.~\eqref{eqn:transition} to the original image $\mathbf{x}_0$ and setting $\alpha_t = 1 - \beta_t$ and $\bar{\alpha}_t = \prod_{i=1}^t \alpha_i$ yields
\begin{equation}
q(\mathbf{x}_t \vert \mathbf{x}_{0}) = \mathcal{N}(\mathbf{x}_t; \sqrt{\bar{\alpha}_t}\mathbf{x}_0, (1-\bar{\alpha}_t)\mathbf{I}). \quad
\end{equation}
Hence $\mathbf{x}_t$ can be written in terms of $\mathbf{x}_0$~\cite{Ho_undated-ak,Nichol2021-ga} 
as 
\begin{equation}
\mathbf{x}_{t}=\sqrt{\bar{\alpha}_t}\mathbf{x}_0 + \sqrt{1 - \bar{\alpha}_t}\mathbf{\epsilon} \quad \text{with} \quad \epsilon \sim \mathcal{N}(0,\mathbf{I}).
\end{equation}
The diffusion model is trained to reverse the forward process by removing noise with each forward pass $p$ through the model, trying to predict $\mathbf{x}_{t-1}$ from the input $\mathbf{x}_{t}$ (see Fig.~\ref{fig:Diffusion_steps} for a visualization of the diffusion process during inference).
Wolleb et al.~\cite{Wolleb2021-xk} used a U-Net model that in each step is trained to predict the probability density function $f(\mathbf{x}_{t})$ from $\mathbf{x}_{t}$ for all $t \in \{1,...,T\}$, where $\mathbf{x}_{t-1}$ serves as the ground truth. With the model parameters denoted by $\theta$, we can then write the reverse process $p_{\theta}$ as
\begin{equation}
p_\theta(\mathbf{x}_{t-1} \vert \mathbf{x}_t) = \mathcal{N}(\mathbf{x}_{t-1};\boldsymbol{\mu}_\theta(\mathbf{x}_t, t),\boldsymbol{\Sigma}_\theta(\mathbf{x}_t,t)).
\end{equation}
Ho et al.~\cite{Ho_undated-ak} derive the formula for the forward pass of the model as
\begin{equation}
\mathbf{x}_{t-1} = \frac{1}{\sqrt{\alpha_{t}}}(\mathbf{x}_{t}\frac{1-\alpha_{t}}{\sqrt{1-\bar{\alpha_{t}}}}\epsilon_{\theta}(\mathbf{x}_{t},t))+\sigma_{t}\textbf{z},
\label{eq:xt-1}
\end{equation}
with $\sigma_{t}$ being the variance scheme the model can learn \cite{Nichol2021-ga}.
Component $\mathbf{z}$ in equation \ref{eq:xt-1} reflects the stochastic sampling process. 
The model is trained with input $\mathbf{x}_{t}=\sqrt{\bar{\alpha_{t}}\mathbf{x}_{0}}+\sqrt{1-\bar{\alpha_{t}}\epsilon}$
to subtract the noise scheme $\epsilon_{\theta}(\mathbf{x}_{t},t)$ from $\mathbf{x}_{t}$ according to equation \ref{eq:xt-1}.

\begin{figure}[t]
\begin{center}
        \rule{0.01\linewidth}{0pt}
        \includegraphics[width=0.83\linewidth]{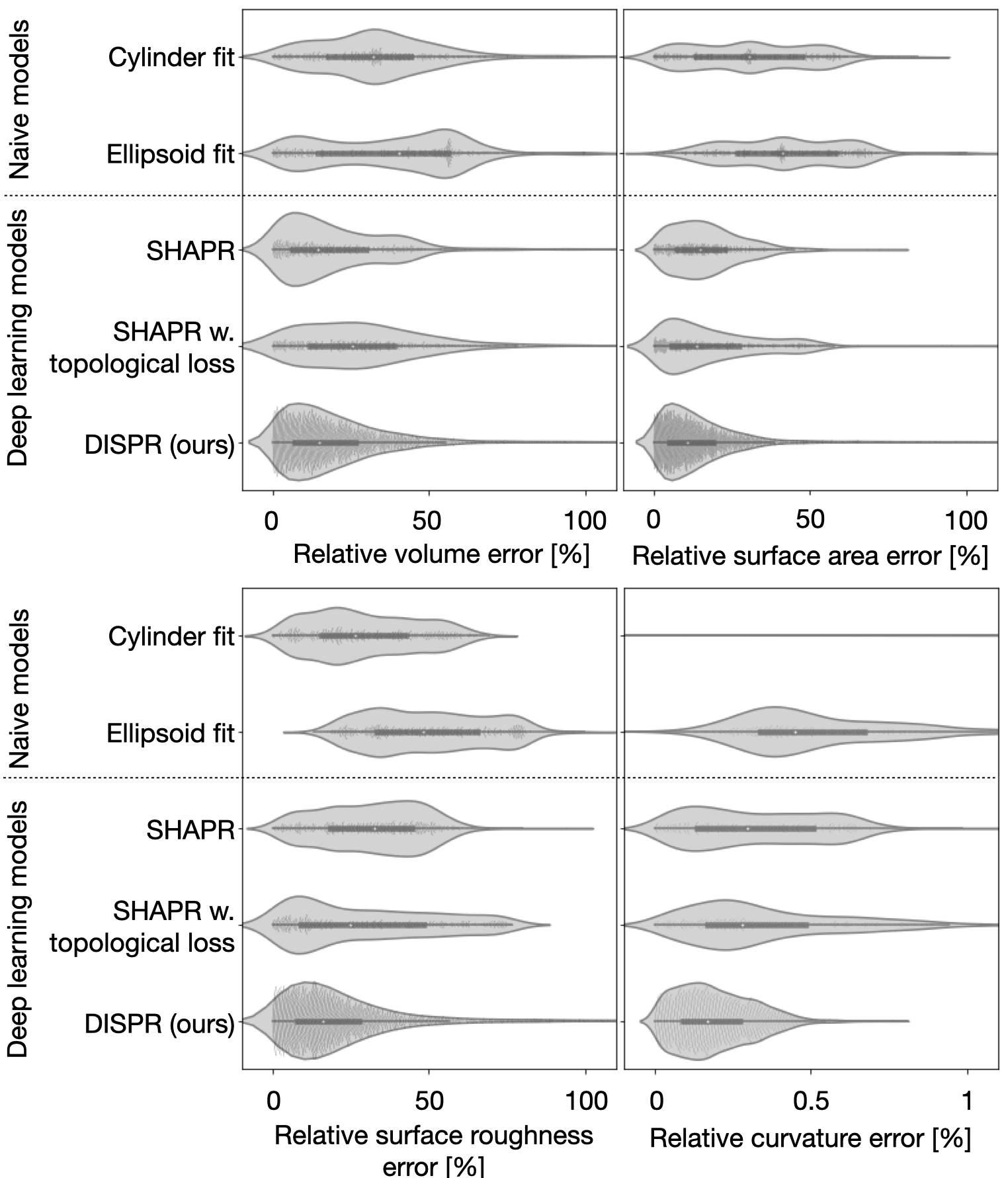}
\end{center}
\caption{With the predictions of \modelname\ we obtain a lower relative volume error as compared to the extrapolations of two naive models, the cylinder and ellipsoid fit, as well as predictions of SHAPR~\cite{Waibel2021-or} and the predictions of SHAPR using the topological loss~\cite{Waibel2022-ng}. 
\modelname\ also outperforms the other models with respect to the surface area error, surface roughness error, and relative surface curvature error. Note that the relative errors of the predictions of \modelname\ contain five times as many datapoints than the others.}
\label{fig:Results}
\end{figure}

\subsection{\modelname}

%Building on previous work by Wolleb et al., who perform a 2D segmentation task, w
We use 2D images to constrain the 3D image generation. In each training and evaluation step the 2D image $b \in \reals^2$ is concatenated to the noisy segmentation mask $\mathbf{x}_{b,t} \in \reals^3$. 
%As opposed to the 2D segmentation task introduced by Wolleb et al., we use a 3D U-Net for the diffusion step to predict 3D shapes.
The groundtruth volume for each input image is $\mathbf{x}_{b,0} \in \reals^3$. 
%Thereby we predict single cell shapes that are realistic reconstructions of the constraint $b$ (see Fig. \ref{fig:Diffusion_SHAPR}). 
%In each forward pass $\mathbf{x}_{b}$ through the model we concatenate the 2D image $b$, while noise is only added to the groundtruth $\mathbf{x}_{b}$. We utilize the linear noise schedule to calculate $\beta_t$ and $\alpha_t$ as proposed by Nichol et al. \cite{Nichol2021-ga}. 
This leads to 
 \begin{equation}
 \mathbf{x}_{b,t-1} = \frac{1}{\sqrt{\alpha_{t}}}\left(\mathbf{x}_{b,t}\frac{1-\alpha_{t}}{\sqrt{1-\bar{\alpha_{t}}}}\epsilon_{\theta}(\mathbf{x}_{b,t}\oplus b,t)\right)+\sigma_{t}\textbf{z}.
 \label{eq:xbt-1}
 \end{equation}
During inference, a noisy image $\mathbf{x}_{b,T}$, is passed $T$ times through the model (Fig.~\ref{fig:Visualization}). 
Because of the stochasticity of $\mathbf{x}_{b,T}$ the diffusion model $f_{\theta}(\mathbf{x}_{T,b}) = p_{\theta}(...(p_{\theta}(\mathbf{x}_{T,b}))$ predicts different outputs $\mathbf{x}_{0}$ after $T=1000$ forward passes through the model, which all represent realistic reconstructions of the 2D image constraint $b$ (see Fig.~\ref{fig:Diffusion_SHAPR}).

\begin{table}[t]
    \begin{center}
    {\small{
    \begin{tabular}{lp{2cm}S[table-format=1.2]SS[round-precision=1]S[table-format=1.3]SS[round-precision=1]}
        \toprule
%        \multicolumn{3}{c}{Red blood cell~($n = 825$)}\\
%        \midrule
        \parbox{2cm}{Relative \\ error} & Model & {Median} & {$\mu\pm\sigma$} \\
        \midrule
        \multirow{3}{*}{Volume} 
        & Cylinder fit  & 0.32 & {0.34 $\pm$ 0.22}\\
        & Ellipsoid fit  & 0.40 & {0.37 $\pm$ 0.23}\\
        & SHAPR  & \bfseries{0.15} & \bfseries{0.20 $\pm$ 0.18}\\
        & Topo SHAPR & 0.26 &  {0.29 $\pm$ 0.27}\\
        & \modelname\ (ours) & \bfseries{0.15} & \bfseries{0.20 $\pm$ 0.20}\\
        \midrule
        \multirow{3}{*}{\parbox{2cm}{Surface \\ area}} 
        & Cylinder fit  & 0.30 & {0.30 $\pm$ 0.19}\\
        & Ellipsoid fit  & 0.41 & {0.42 $\pm$ 0.19}\\
        & SHAPR  & 0.15 & {0.16 $\pm$ 0.11}\\
        & Topo SHAPR & 0.14 & {0.18 $\pm$ 0.16}\\ 
        & \modelname\ (ours) & \bfseries{0.11} & \bfseries{0.14 $\pm$ 0.15}\\ 
        \midrule
        \multirow{3}{*}{\parbox{2cm}{Surface \\ roughness}}      
        & Cylinder fit  & 0.26 & {0.29 $\pm$ 0.17}\\
        & Ellipsoid fit  & 0.48 & {0.49 $\pm$ 0.19}\\
        & SHAPR  & 0.32 & {0.31 $\pm$ 0.16}\\
        & Topo SHAPR &  0.25 &  {0.29 $\pm$ 0.29} \\
        & \modelname\ (ours) & \bfseries{0.15} & \bfseries{0.23 $\pm$ 0.27}\\ 
        \midrule
        \multirow{3}{*}{Curvature} 
        & Cylinder fit  & 4.79 & {4.76 $\pm$ 1.36}\\
        & Ellipsoid fit  & 0.38 & {0.39 $\pm$ 0.17}\\
        & SHAPR  & 0.30 & {0.32 $\pm$ 0.21}\\
        & Topo SHAPR &  0.28 &  {0.34 $\pm$ 0.24}\\ 
        & \modelname\ (ours) & \bfseries{0.17} & \bfseries{0.19 $\pm$ 0.12}\\ 
         \bottomrule
  \end{tabular}
}}
\end{center}
\caption{
        Median, mean~($\mu$), and standard deviation~($\sigma$) of 
        relative errors~(lower values are better;
        best result is marked in \textbf{bold}).}
    \label{tab:Results}
\end{table}

\subsection{Dataset}
\label{subsec:dataset}

We use a publicly-available red blood cell dataset published by Simionato et al.~\cite{Simionato21a}.
It consists of $825$ 3D images of red blood cells recorded with a confocal microscope. 
All cells are available as segmented 3D images~\cite{Waibel2021-or,dominik_waibel_2022_7031924}. Each image is contained in a $64 \times 64 \times 64$ voxel grid.
Each cell is assigned to one of the following six classes, with $n$ beeing the number of images in each class:
\begin{inparaenum}[(i)]
    \item SDE shapes ($n=602$),
    \item cell clusters ($n=69$),
    \item multilobates ($n=12$),
    \item keratocytes ($n=31$),
    \item knizocytes ($n=23$), or
    \item acanthocytes ($n=88$).
\end{inparaenum}
Note that Spherocytes, stomatocytes, discocytes, and echinocytes are combined into the SDE shapes class, which is characterized by the stomatocyte--discocyte--echinocyte transformation~\cite{Simionato21a}.
%
%The dataset is imbalanced, with the number of images in each class being
%
%$602$ for SDE shapes~($93$ spherocytes, $41$ stomatocytes, $176$ discocytes, and $292$ echinocytes), 
%$69$ cell clusters,
% $12$ multilobates,
%$31$ keratocytes,
%$23$ knizocytes, and
%$88$ acanthocytes. 
%
The 2D images contained in the provided dataset have been extracted from the central slide of each 3D image and segmented by thresholding, thus resulting in a (64,64) pixel size.

% \begin{figure}[!ht]
% \begin{center}
%     \fbox{%\rule{0pt}{2in} 
%         \rule{0.01\linewidth}{0pt}
%         \includegraphics[width=0.8\linewidth]{Figures/WACV_DiffusionSHAPR_Figures.005.png}}
% \end{center}
%     \caption{Morphological features extracted from the predictons of \modelname\ reflect the same celltype-specific distribution as the features extracted from the 3D groundtruth. This indicates that \modelname has learned class-specific morphological features and motivates a morphological feature based classification approach.}
%     \label{fig:feature_comparison}
% \end{figure}

\begin{figure}[t]
\begin{center}
    \fbox{%\rule{0pt}{2in} 
        \rule{0.01\linewidth}{0pt}
        \includegraphics[width=0.95\linewidth]{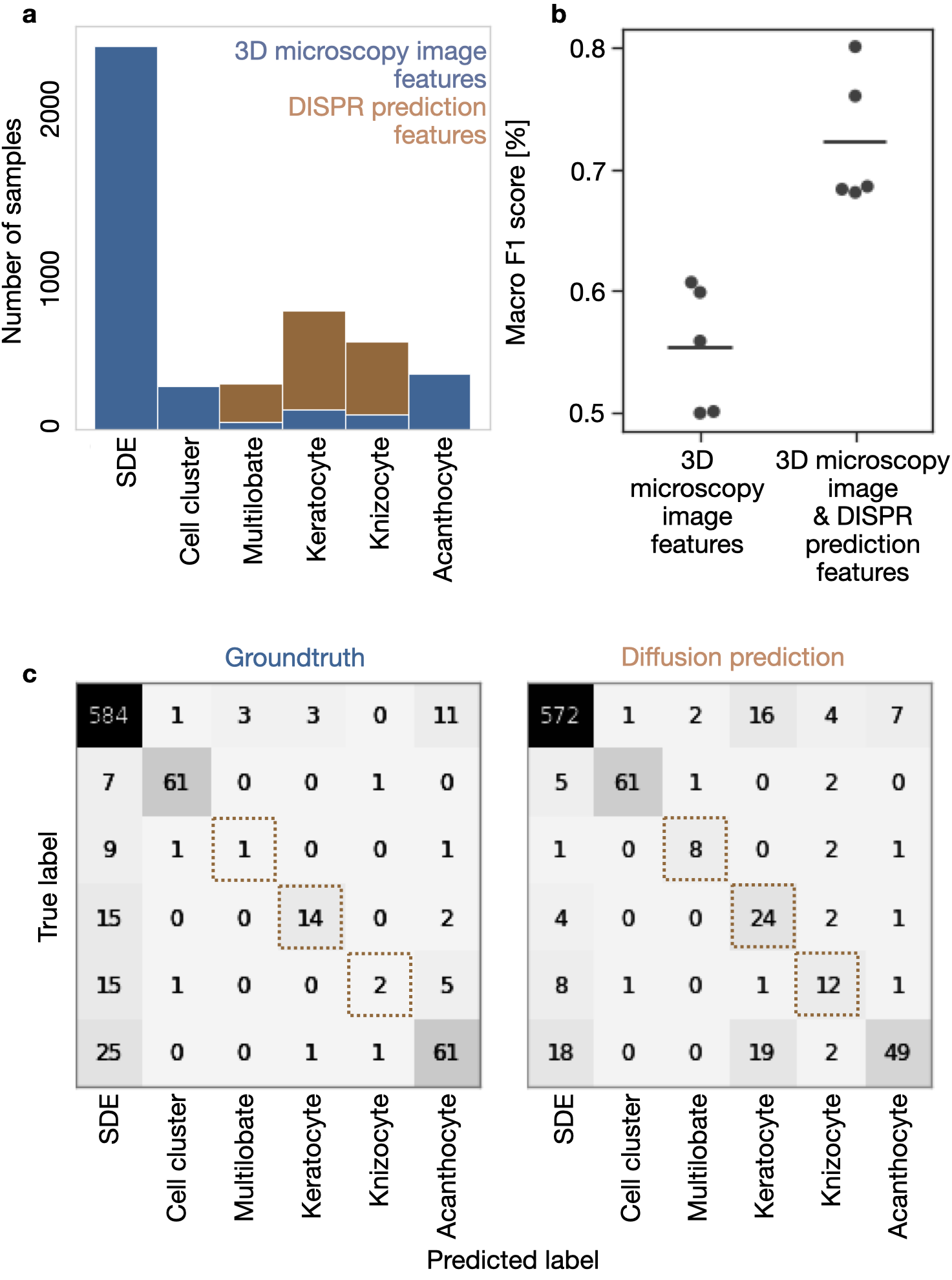}}
\end{center}
\caption{Using \modelname\ to oversample training data improves a feature-based random forest classification of single red blood cells. (a) For the training dataset, containing 128 manually extracted morphological features of the 3D groundtruth, we enhance the three smallest classes (multilobate, keratocyte, knizocyte) with features extracted from the \modelname's predictions, reducing class imbalance.
(b) We improve the macro F1 score to $F1_{\text{macro}} = 72.2\pm4.9\%$, from $F1_{\text{macro}}$ = $55.2\pm4.6\%$, with the largest improvements obtained for minority classes (c).}
\label{fig:random_forest}
\end{figure}

\section{Experiments}
%\subsection{Training}

We split the dataset into five folds with a $80\%$/$20\%$ train/test split, ensuring that each image is contained in the test set exactly once. 
We trained five separate models, one for each split for 85000 steps with $T=1000$, and the hyper-parameters provided by Wolleb et al.~\cite{Wolleb2021-xk}. 
We trained and evaluated the model on a compute cluster containing Tesla V100 GPUs with 16GB memory. 
Due to GPU memory constraints we had to run the models with a batch-size of one. We used a learning rate of $10^{-4}$, with no weight decay, a linear noise scheduler, and the Adam optimizer. 
% \cite{Kingma2014-cx}. 
The number of channels in the first layer were 32, and the layers of our multihead attention layers had a resolution of 16. We used a uniform sample scheduler for the diffusion steps.
Each model was trained for 8500 steps which took roughly 60 hours, summing up to 300 hours of training time for five models. During inference the model took on average 4:45 minutes for each prediction, summing up to approximately 330 hours for 4125 predictions, as we predict five 3D cell shapes for each 2D input image. 

%\subsection{Evaluation}
We evaluate each of the five trained diffusion models on every image~$b$ in the test set five times to obtain five shape predictions.
As each model is tested on 165 2D images, this adds up to $165\cdot 25 = 4125$ predicted cell shapes. 
Due to the stochasticity of the diffusion model, variations between the predicted shapes occur and each predicted cell is unique (see Fig. \ref{fig:Visualization}).
To quantify the performance of our model, we compare biologically relevant features such as the volume, surface area, surface roughness, and convexity to the groundtruth to reconstructions obtained from SHAPR~\cite{Waibel2021-or} or its topology-aware variant~\cite{Waibel2022-ng}; to date, these models constitute the state-of-the-art for this shape reconstruction task.
As additional comparisons, we use two naive baseline models, a cylinder and ellipsoid fit, where we extrapolate the 3D shape from the 2D segmentation mask, similar to Waibel et al.~\cite{Waibel2021-or}. 
We calculate the mean of the major and the minor axis of an ellipse fitted to the 2D segmentation outline and use this as the cylinder's extent in the third dimension and as the third axis for the ellipsoid fit, respectively. 

%\subsection{Morphological feature based classification}

\section{Results}

In our experiments we find high visual morphological consistency between the five predictions and also a high similarity to the groundtruth 3D shape.
For individual features, such as the volume, surface area, surface roughness and curvature our approach outperforms the four competing approaches~\cite{Waibel2021-or,Waibel2022-ng} (see Fig.~\ref{fig:Results} and 
Table~\ref{tab:Results}) in terms of relative error to the groundtruth.

%\subsection{Morphological feature similarity}
%For each of the nine  cell types, which were sorted in six classes, contained in the dataset, the feature distributions of the groundtruth features and predicted features exhibit high similarity,
%(see Fig.~\ref{fig:feature_comparison}), 
%indicating that \modelname\ has learned to extract class-specific information in the 2D images and predicts class-specific morphological features. 

To test if DISPR based training data generation improves single cell classification, we extract 128 morphological features from each single cell, similar to Waibel et al.~\cite{Waibel2021-or}.
Using random forest classifiers, we predict the respective class of each cell based on the extracted morphological features.
Five random forest models are trained to classify each cell in one of the six classes suggested by Simionato et al. \cite{Simionato21a} in a round-robin fashion, splitting the dataset with a $80\%$/$20\%$ train/test split, ensuring each image is contained in the test set exactly once. The random forest models are trained with 1000 estimators and a depth of 10.

% The potential of \modelname\ to generate 3D cell shapes with class specific morphological properties was demonstrated by this feature based classification task. A baseline was established on the features extracted from the segmented 3D microscopy images used as the groundtruth for training \modelname.  

To oversample the three smallest classes, we predict five 3D cell shapes for each 2D image of the three smallest classes contained in the dataset (multilobates, keratocytes, and knizocytes) (see Fig. \ref{fig:random_forest}a). In a second step, we add features extracted from \modelname's predictions to the respective training sets, if the same cell was already contained in the training dataset. Thereby we ensure that no overlap between the train and test set occurs.
By enriching the groundtruth features with features obtained from the predictions of \modelname, the classification performance significantly increased ($F1_{\text{macro}} = 72.2\pm4.9\%$, $F1_{\text{weighted}}$ = $88.6\pm1.1\%$,  mean$\pm$sdev, $n=5$ cross-validation runs) as compared to using groundtruth microscopy image features only ($F1_{\text{macro}}$ = $55.2\pm4.6\%$, $F1_{\text{weighted}}$ = $85.7\pm2.1\%$ (see Fig \ref{fig:random_forest}b)) in a fivefold cross-validation (see Fig.~\ref{fig:random_forest}b).
The largest improvements were found for those classes to which we added features, especially for the keratocytes, the number of correctly classified cells increased from 14 to 24 and for the multilobate cells from 1 to 8 correct classifications (see Fig \ref{fig:random_forest}c).

\section{Discussion}
% \noindent
% \textbf{\modelname\ reconstructs realistic 3D cell shapes}
We demonstrate that \modelname reconstructs 3D shapes from 2D microscopy images by comparing relevant morphological features.
% The inference of \modelname\ is visualized in Fig. \ref{fig:Diffusion_steps}, where the model reconstructs the 3D shape of a stromatocyte from Gaussian noise. This figure suggests that the model accumulates it's transformative performance in a limited number of steps, however we point out that this is only a visualization, and the model might perform relevant reconstruction steps during other steps $t$, which might not be visible for us behind the noise. 
Class-specific morphological feature distributions from the ground truth dataset are captured in the model distribution and suggests that our approach is capable of data augmentation.

The lower relative surface roughness and curvature error further suggest improvements in predicting morphological properties compared to other state of the art models. 
This also gives credence to our visual impression, that our model is able to predict 3D shapes that are more realistic.
% \noindent
% \textbf{Morphological feature based classification with \modelname's predictions.}
Additionally, we show that the morphological features do not only appear more realistic, but \modelname\ can be utilized for generating training data, used in a classification task, leading to significant improvements especially on minority classes, emphasising the utility of \modelname\ for generating data.
\modelname, is not restricted to single cells, and may be applied to other reconstruction tasks as well.

We acknowledge that DISPR's inference is magnitudes less efficient as compared to autoencoder models, such as SHAPR \cite{Waibel2021-or,Waibel2022-ng} for 3D shape prediction, where each prediction takes a few seconds, as compared to 4:45minutes using DISPR.

% \noindent
% \textbf{Future directions.}
One future direction could be to consider geometrical and topological information in the model's regularization to improve training efficiency and model performance. 
% We speculate that the inclusion of additional inductive biases such equivariance or even invariance to certain transformations may be highly beneficial for biomedical reconstruction tasks, where the \emph{shape} of an object matters much more than its orientation with respect to space~(note that this is in stark contrast to real-world images, where relative positions of objects in an image may carry a large amount of semantic meaning).
% In the trade-off between throughput and the resolution of our 3D shape prediction, we find that the increased resolution comes at the expense of computational resources, due to the fact that each prediction requires $T=1000$ denoising steps. This is an inherent property of diffusion models, and a natural line of work is to investigate if faster inference times can be achieved.
% Utilizing a diffusion model's ability to predict a \emph{distribution} of single cell shapes of the same 2D microscopic image, we envision to bridge the gap of uncertainty estimation, for example, similar to using Monte Carlo Dropout. 
Exploiting the distributional nature of predicted shapes could lead to additional studies of their structure; cells whose reconstructions are highly ambiguous could be detected by a multi-modal shape reconstruction.
%Moreover, it would be relevant to investigate how diffusion models behave under adversarial attacks or as a generator of adversarial samples for attacking another model. 
Finally, to reduce computational cost during inference, which is one of the main drawbacks of diffusion models, super-resolution models paired with diffusion models \cite{ho2022cascaded} might prove useful in 3D applications. 
% Another approach are diffusion models that are applied on the latent space of pretrained autoencoders, saving computational resources \cite{rombach2022high}.
In the biomedical domain denoising images is a recurring task, which might be solvable with diffusion models, as their training objective itself is the denoising of images.

\subsection{Data and code availability}
The dataset is available at \cite{dominik_waibel_2022_7031924}: doi: 10.5281/zenodo.7031924 

\noindent
The code is available at: https://github.com/marrlab/DISPR

\subsection{Compliance with Ethical Standards}
No ethical approval was required for this study.

\clearpage

\subsection{Acknowledgements} We thank Julia Wolleb (Basel) for the inspiration, Melanie Schulz and Daniel Lang (Munich), and Erez Yosef (Tel Aviv) for discussing ideas and feedback to this manuscript. We also gratefully acknowledge the use of Helmholtz Munich High-Performance Computing Cluster for model training. Using the ML $CO_2$ calculator \cite{lacoste2019quantifying} we estimate 81.65kg of $CO_2$ emitted, considering only experiments. 

\subsection{Funding} Carsten Marr received funding from the European Research Council (ERC) under the European Union’s Horizon 2020 Research and Innovation Programme (Grant Agreement 866411).

% \subsection{Author contributions} DW implemented code, conducted experiments, wrote the manuscript and created figures. RG, BR, and CM supervised the study. All authors have read and approved the manuscript.

% References should be produced using the bibtex program from suitable
% BiBTeX files (here: strings, refs, manuals). The IEEEbib.bst bibliography
% style file from IEEE produces unsorted bibliography list.
% ------------------------------------------------------------------------- 
\bibliographystyle{IEEEbib}
\bibliography{strings,refs}

\begin{thebibliography}{10}

\bibitem{diez2010shape}
Monica Diez-Silva, Ming Dao, Jongyoon Han, Chwee-Teck Lim, and Subra Suresh,
\newblock ``Shape and biomechanical characteristics of human red blood cells in
  health and disease,''
\newblock {\em MRS bulletin}, vol. 35, no. 5, pp. 382--388, 2010.

\bibitem{Waibel2021-or}
Dominik~J.E. Waibel, Niklas Kiermeyer, Scott Atwell, Ario Sadafi, Matthias
  Meier, and Carsten Marr,
\newblock ``Shapr predicts 3d cell shapes from 2d microscopic images,''
\newblock {\em iScience}, vol. 25, no. 11, pp. 105298, 2022.

\bibitem{Waibel2022-ng}
Dominik J.~E. Waibel, Scott Atwell, Matthias Meier, Carsten Marr, and Bastian
  Rieck,
\newblock ``Capturing shape information with multi-scale topological loss
  terms for 3d reconstruction,''
\newblock in {\em Medical Image Computing and Computer Assisted Intervention --
  MICCAI 2022}, Linwei Wang, Qi~Dou, P.~Thomas Fletcher, Stefanie Speidel, and
  Shuo Li, Eds., Cham, 2022, pp. 150--159, Springer Nature Switzerland.

\bibitem{Gkioxari2019-ba}
Georgia Gkioxari, Jitendra Malik, and Justin Johnson,
\newblock ``Mesh {R-CNN},''
\newblock in {\em Proceedings of the IEEE/CVF International Conference on
  Computer Vision~(ICCV)}, 2019.

\bibitem{Wang2018-qg}
Nanyang Wang, Yinda Zhang, Zhuwen Li, Yanwei Fu, Wei Liu, and Yu-Gang Jiang,
\newblock ``\texttt{Pixel2mesh}: Generating {3D} mesh models from single {RGB}
  images,''
\newblock in {\em Proceedings of the European Conference on Computer
  Vision~({ECCV})}, 2018, pp. 52--67.

\bibitem{Choy2016-fa}
Christopher~B Choy, Danfei Xu, Junyoung Gwak, Kevin Chen, and Silvio Savarese,
\newblock ``{3D-R2N2}: A unified approach for single and multi-view {3D} object
  reconstruction,''
\newblock in {\em Proceedings of the European Conference on Computer
  Vision~(ECCV)}, 2016, pp. 628--644.

\bibitem{watson2022novel}
Daniel Watson, William Chan, Ricardo Martin-Brualla, Jonathan Ho, Andrea
  Tagliasacchi, and Mohammad Norouzi,
\newblock ``Novel view synthesis with diffusion models,''
\newblock {\em arXiv preprint arXiv:2210.04628}, 2022.

\bibitem{Fan2017-uf}
Haoqiang Fan, Hao Su, and Leonidas~J. Guibas,
\newblock ``A point set generation network for {3D} object reconstruction from
  a single image,''
\newblock in {\em Proceedings of the IEEE Conference on Computer Vision and
  Pattern Recognition (CVPR)}, 2017.

\bibitem{Saharia2022-xs}
Chitwan Saharia, William Chan, Saurabh Saxena, Lala Li, Jay Whang, Emily
  Denton, Seyed Kamyar~Seyed Ghasemipour, Burcu~Karagol Ayan, S~Sara Mahdavi,
  Rapha~Gontijo Lopes, et~al.,
\newblock ``Photorealistic text-to-image diffusion models with deep language
  understanding,''
\newblock {\em arXiv preprint arXiv:2205.11487}, 2022.

\bibitem{Nichol2021-ga}
Alexander~Quinn Nichol and Prafulla Dhariwal,
\newblock ``Improved denoising diffusion probabilistic models,''
\newblock in {\em Proceedings of the 38th International Conference on Machine
  Learning}, Marina Meila and Tong Zhang, Eds. 2021, vol. 139 of {\em
  Proceedings of Machine Learning Research}, pp. 8162--8171, PMLR.

\bibitem{Song2020-zk}
Jiaming Song, Chenlin Meng, and Stefano Ermon,
\newblock ``Denoising diffusion implicit models,''
\newblock {\em arXiv preprint arXiv:2010.02502}, 2020.

\bibitem{Dhariwal_undated-xt}
{Dhariwal} and {Nichol},
\newblock ``Diffusion models beat gans on image synthesis,''
\newblock {\em Adv. Neural Inf. Process. Syst.}

\bibitem{Wolleb2021-xk}
Julia Wolleb, Robin Sandk{\"u}hler, Florentin Bieder, Philippe Valmaggia, and
  Philippe~C Cattin,
\newblock ``Diffusion models for implicit image segmentation ensembles,''
\newblock {\em arXiv preprint arXiv:2112.03145}, 2021.

\bibitem{Wolleb2022-sq}
Julia Wolleb, Florentin Bieder, Robin Sandk{\"u}hler, and Philippe~C Cattin,
\newblock ``Diffusion models for medical anomaly detection,''
\newblock {\em arXiv preprint arXiv:2203.04306}, 2022.

\bibitem{khader2022medical}
Firas Khader, Gustav Mueller-Franzes, Soroosh~Tayebi Arasteh, Tianyu Han,
  Christoph Haarburger, Maximilian Schulze-Hagen, Philipp Schad, Sandy
  Engelhardt, Bettina Baessler, Sebastian Foersch, et~al.,
\newblock ``Medical diffusion--denoising diffusion probabilistic models for 3d
  medical image generation,''
\newblock {\em arXiv preprint arXiv:2211.03364}, 2022.

\bibitem{ozbey2022unsupervised}
Muzaffer {\"O}zbey, Salman~UH Dar, Hasan~A Bedel, Onat Dalmaz, {\c{S}}aban
  {\"O}zturk, Alper G{\"u}ng{\"o}r, and Tolga {\c{C}}ukur,
\newblock ``Unsupervised medical image translation with adversarial diffusion
  models,''
\newblock {\em arXiv preprint arXiv:2207.08208}, 2022.

\bibitem{ali2022spot}
Hazrat Ali, Shafaq Murad, and Zubair Shah,
\newblock ``Spot the fake lungs: Generating synthetic medical images using
  neural diffusion models,''
\newblock {\em arXiv preprint arXiv:2211.00902}, 2022.

\bibitem{yoon2022sadm}
Jee~Seok Yoon, Chenghao Zhang, Heung-Il Suk, Jia Guo, and Xiaoxiao Li,
\newblock ``Sadm: Sequence-aware diffusion model for longitudinal medical image
  generation,''
\newblock {\em arXiv preprint arXiv:2212.08228}, 2022.

\bibitem{kim2022diffusion}
Boah Kim and Jong~Chul Ye,
\newblock ``Diffusion deformable model for 4d temporal medical image
  generation,''
\newblock in {\em Medical Image Computing and Computer Assisted
  Intervention--MICCAI 2022: 25th International Conference, Singapore,
  September 18--22, 2022, Proceedings, Part I}. Springer, 2022, pp. 539--548.

\bibitem{Ho_undated-ak}
{Ho}, {Jain}, and {Abbeel},
\newblock ``Denoising diffusion probabilistic models,''
\newblock {\em Adv. Neural Inf. Process. Syst.}

\bibitem{Simionato21a}
Greta Simionato, Konrad Hinkelmann, Revaz Chachanidze, Paola Bianchi, Elisa
  Fermo, Richard van Wijk, Marc Leonetti, Christian Wagner, Lars Kaestner, and
  Stephan Quint,
\newblock ``Red blood cell phenotyping from {3D} confocal images using
  artificial neural networks,''
\newblock {\em PLOS Computational Biology}, vol. 17, no. 5, pp. 1--17, 2021.

\bibitem{dominik_waibel_2022_7031924}
Dominik Waibel, Niklas Kiermeyer, Scott Atwell, Bastian Rieck, Matthias Meier,
  and Carsten Marr,
\newblock ``{Datasets for 3D shape reconstruction from 2D microscopy images},''
  Sept. 2022.

\bibitem{ho2022cascaded}
Jonathan Ho, Chitwan Saharia, William Chan, David~J Fleet, Mohammad Norouzi,
  and Tim Salimans,
\newblock ``Cascaded diffusion models for high fidelity image generation.,''
\newblock {\em J. Mach. Learn. Res.}, vol. 23, pp. 47--1, 2022.

\bibitem{lacoste2019quantifying}
Alexandre Lacoste, Alexandra Luccioni, Victor Schmidt, and Thomas Dandres,
\newblock ``Quantifying the carbon emissions of machine learning,''
\newblock {\em arXiv preprint arXiv:1910.09700}, 2019.

\end{thebibliography}

\end{document}